\documentclass[11pt]{article}
\usepackage{placeins}
\usepackage[final]{acl}

\usepackage{times}
\usepackage{latexsym}

\usepackage[T1]{fontenc}

\usepackage[utf8]{inputenc}

\usepackage{microtype}

\usepackage{inconsolata}

\usepackage{graphicx}
\usepackage[inline]{enumitem} 
\setlist[itemize]{leftmargin=*, nosep}
\setlist[enumerate]{leftmargin=*,nosep,label=(\arabic*)}
\usepackage[font=small,labelfont=bf]{caption}
\usepackage{tikz}
\usepackage{adjustbox}

\graphicspath{{Figures/}}

\usepackage{listings}
\usepackage{adjustbox} 
\usepackage{booktabs}
\usepackage{multirow} 
\usepackage{array}   

\usepackage{pifont}

\usepackage{amsmath}  

\usepackage{amssymb}

\usepackage{subcaption} 

\usepackage{acronym}
\acrodef{AI}{artificial intelligence}
\acrodef{LLM}{large language model}
\acrodef{SOTA}{state-of-the-art}
\acrodef{TED}{tree edit distance}
\acrodef{SMR}{sequence matcher ratio}
\acrodef{LD}{Levenshtein distance}
\acrodef{SR}{symbolic regression}
\acrodef{MWP}{math word problem}

\usepackage{algorithm}
\usepackage{algpseudocode}

\usepackage[most]{tcolorbox}

\usepackage{xcolor}

\title{SciText2Eq: Assessing LLMs for Explainable Equation Generation for Scientific Creativity}

\author{
\textbf{Yifan Mo\textsuperscript{1}},
\textbf{Xiao Fu\textsuperscript{1}},
\textbf{Yue Su\textsuperscript{1}},
\textbf{Qingyu Meng\textsuperscript{1}},
\textbf{Koen Hindriks\textsuperscript{1}},
\textbf{Qingzhi Liu\textsuperscript{2}},
\textbf{Jiahuan Pei\textsuperscript{1}\thanks{Corresponding author.}}
\\
\\
\textsuperscript{1}Vrije Universiteit Amsterdam, Amsterdam, Netherlands \\
\textsuperscript{2}Wageningen University \& Research, Netherlands \\
{\small
\texttt{\{y.mo, y.su, q.meng, k.v.hindriks, j.pei2\}@vu.nl},  \texttt{ceciliafu2000@gmail.com}, \texttt{qingzhi.liu@wur.nl}
}
}

\begin{document}
\maketitle
\begin{abstract}
This work investigates the ability of large language models (LLMs) to generate mathematical equations from scientific texts. 
Prior work faces challenges in unstructured grounding, multi-equation dependency, and human-aligned evaluation.
To this end, we construct a dataset of AI research papers, pairing contextual passages with ground-truth equations and variable descriptions. 
We develop an explainable equation generation workflow and evaluate it across diverse open- and closed-source LLM backbones.
We introduce an evaluation protocol combining automatic metrics, LLM-based rubrics, and human judgments to assess accuracy, explainability, and human-LLM alignment.
Results indicate that LLMs perform moderately on lexical- and syntactic-based similarity, while struggling with semantic accuracy. 
Comparisons between LLM-based evaluations and human judgments reveal limited alignment, highlighting challenges in using LLMs to assess equation quality. 
These findings offer insights for improving equation generation models and developing more reliable evaluation methods for scientific text.
We provide code and data for reproducibility.~\footnote{\url{https://github.com/YifanMo727/SciText2Eq}}
\end{abstract}

\section{Introduction}
Mathematical equations underpin scientific knowledge representation and reasoning~\cite{bais2025equations}. Recent research include
equation recognition~\cite{zhou2022end}, equation retrieval~\citep{wang2021scientific,vemuganti2025advancing}, mathematical language understanding~\citep{peng2021mathbert,scarlatos2023tree}, equation reasoning~\citep{meadows2025controlling,yu2025formalmath}.
As scientific literature grows denser, generating and explaining equations in natural language has become central to scientific creativity~\cite{yasunaga2019topiceq}.

\begin{figure}[t]
    \centering
\includegraphics[width=\linewidth,
        trim=10 10 20 5,
        clip]{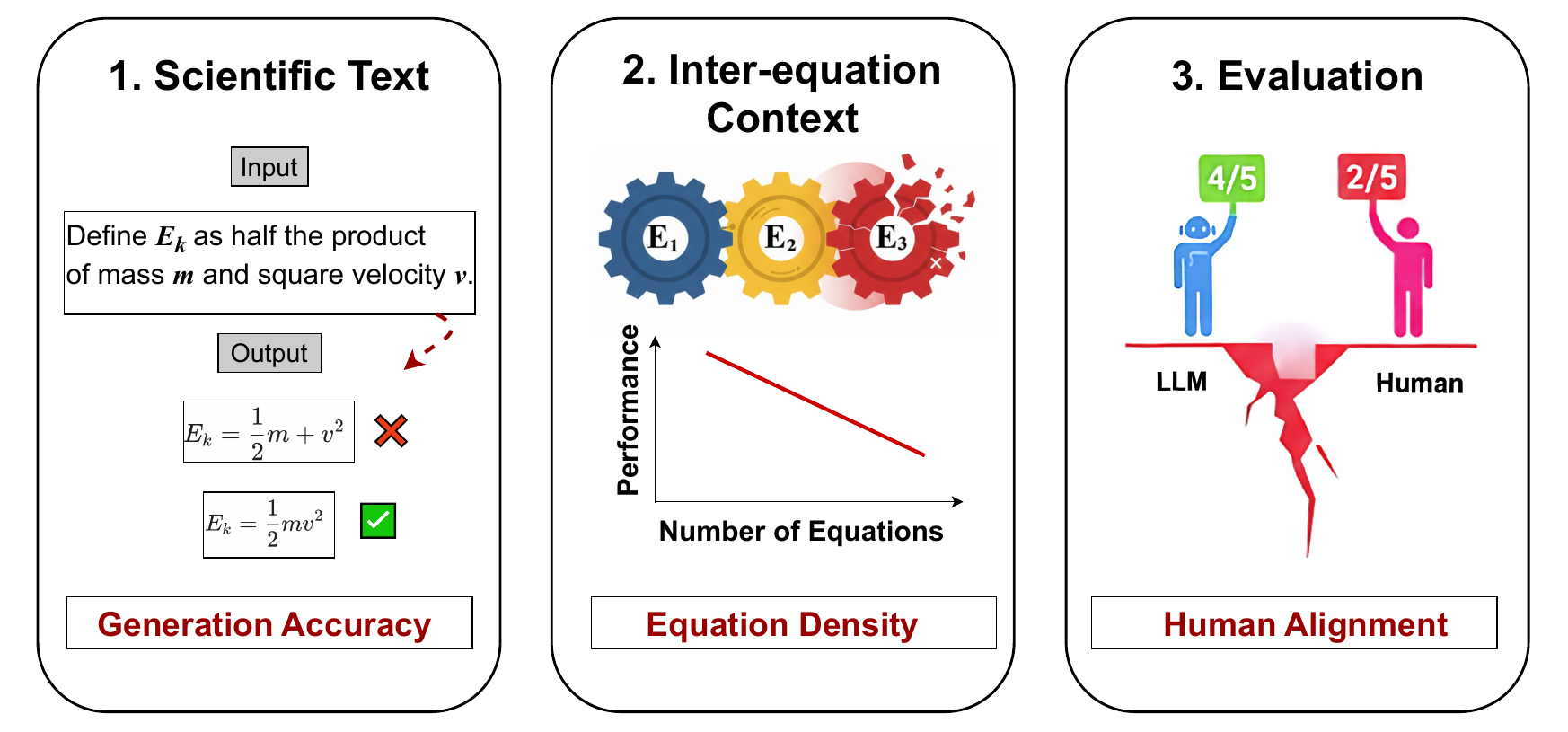}
    \caption{Three key challenges in generating equations from unstructured text. 
    }
    \label{fig:motivation}
\end{figure} 

A key line of related work is equation discovery, or \ac{SR}, which recovers mathematical expressions describing relationships among observed variables~\citep{dong2025recent}. 
Recent approaches leverage LLMs to propose and refine candidate equations using programming~\cite{shojaee2024llm}, reasoning~\citep{wang2025drsr}, and benchmarking~\citep{shojaee2025llm}. 
In contrast, equation generation focuses on producing expressions conditioned on textual context. 
Prior work includes extracting single equations from unstructured narrative text~\citep{zhang2023expression,liang2023let,zong2023solving}. 
TopicEq jointly models topics and equations to align expressions with scientific text~\citep{yasunaga2019topiceq}, yet its dataset is not publicly available and contexts remain relatively simple. 

In real scientific texts, equations appear in unstructured narratives, requiring explanation, inter-equation reasoning, and semantic fidelity.
There are three major challenges:
\begin{enumerate*}
    \item accurately generating mathematical expressions from scientific texts where equations are interleaved with natural language descriptions; 
    \item handling multiple, non-independent equations within the same document and understanding how equation density and inter-equation context affect generation accuracy; 
    \item evaluating generated equations in a comprehensive way and aligning \ac{LLM}-based judgments with human assessments, while maintaining explainability of the process. 
\end{enumerate*}
These challenges reveal a gap in generating publication-quality scientific equations (\autoref{fig:motivation}).

In this work, we propose \textbf{SciText2Eq}, a comprehensive workflow for scientific discovery by \acp{LLM} in real-world scientific text, designed to systematically diagnose how well \acp{LLM} can reconstruct and reason about mathematical expressions for scientific creativity. We construct a dedicated dataset of \acs{AI} research papers, selecting sections rich in equations to provide realistic multi-equation contexts. Equations and their corresponding variable descriptions are sequentially generated by \acp{LLM} from the constructed dataset, based on all context and equations preceding the target equation within each paper. Generated equations are evaluated using a multi-level protocol that combines: 
\begin{enumerate*}
    \item automatic metrics measuring lexical and syntactic similarity;
    \item rubric-based \ac{LLM}-as-judge evaluation assessing validity, coverage, clarity, appropriateness, and equivalence of both equations and accompanying descriptions; and
    \item human evaluation to quantify alignment between \ac{LLM} and expert judgment.
\end{enumerate*}

The contributions of this work are threefold:

\begin{itemize}
  
\item A workflow for scientific equation generation using \acp{LLM}, providing both variable-level explanations and interpretable evaluation rationales.
\item A comprehensive multi-level evaluation framework, combining automatic metrics, LLM-based rubric metrics, and human judgment, along with an extensible benchmark dataset.
\item An empirical study of equation density, revealing its impact on equation generation performance.
\end{itemize}

\section{Related work}

\subsection{Equation Discovery and Generation}
Equation discovery, also know as \ac{SR}, aims to recover mathematical expressions from observational data or variables. Traditional SR methods include evolutionary algorithms~\citep{koza1994genetic} and gradient-based optimization~\citep{petersen2021deep}, while more recently, \acp{LLM} have been applied to these tasks. As shown in Appendix~\ref{sec:most_related_work}, most approaches use variable pairs as primary input, with general SR methods performing equation discovery without additional guidance~\cite{Merler2024icsr}, and domain-specific methods incorporating structured information such as scientific priors~\citep{wang2025drsr, song2025llmfe}, symbol libraries~\cite{du2024large}, or brief natural language descriptions of tasks to constrain the search space and guide equation generation~\citep{shojaee2024llm,shojaee2025llm}. 

In contrast, equation generation produces mathematical expressions from textual context rather than variable pairs. Early work uses natural language \acp{MWP} from public datasets~\citep{zhang2023expression,liang2023let,zong2023solving}, such as MathQA~\cite{amini2019mathqa} and Math23K~\cite{wang2017deep}, which mainly contain synthetic arithmetic or algebra problems where text maps directly to equations, requiring little domain knowledge or reasoning. Some approaches also generate equations by modeling the semantic roles of numbers and operators within the problem context~\cite{chiang2019semantically}. Recent work considers real scientific text by pairing each equation with surrounding sentences~\citep{yasunaga2019topiceq}, but it focuses on local context and does not fully capture dependencies across broader sections of a paper.

In summary, while existing work has advanced equation discovery and generation, most methods either focus on structured or synthetic inputs~\citep{Merler2024icsr,shojaee2024llm} or assume that equations in real scientific texts depend primarily on local context~\cite{yasunaga2019topiceq}. This leaves a gap in generating equations from complex, long-form scientific narratives that require integrating global context, which is the focus of our work.

\subsection{Equation Density}

Most benchmarks for equation discovery focus on \ac{SR} tasks with short, independent variable pairs, while domain-specific methods may use priors, symbol libraries, or brief task descriptions~\citep{wang2025drsr,shojaee2025llm,du2024large}, yet equations remain isolated without broader context. Similarly, prior work on equation generation often considers independent problems or short text snippets, generating each equation separately~\cite{zhang2023expression,liang2023let} or using only limited local context~\cite{yasunaga2019topiceq}. Real scientific papers, however, contain multiple interdependent equations within long narratives. 

Overlooking broader context and inter-equation dependencies can limit a model’s ability to capture semantic structure. To better reflect generation performance, our dataset preserves full narrative context, providing all preceding text and equations for each target formula, which enables analysis of how equation density affects generation quality.

\subsection{Evaluation and Human Alignment}
Prior work on equation discovery and generation has largely relied on automated metrics. In symbolic regression, common metrics such as NMSE~\citep{wang2025drsr,shojaee2024llm}, NRMSE~\cite{du2024large}, R$^2$~\cite{Merler2024icsr}, quantify how closely generated equations match observed data~(\autoref{tab:sr_comparison}). For equation generation from natural language, evaluation often focuses on numerical or structural equivalence. Models are assessed on whether generated equations yield the correct numeric results~\citep{zhang2023expression,liang2023let} or on language-model-based metrics such as perplexity and syntax error rate for LaTeX sequences~\citep{yasunaga2019topiceq}. Some approaches incorporate \acp{LLM} as evaluators~\citep{shojaee2025llm}, but they generally assess only one dimension at a time and seldom consider semantic consistency, contextual appropriateness, or clarity of variable roles.

Human evaluation has been relatively limited in prior work. In equation discovery, few studies compare LLM-generated outputs with human judgment~\citep{shojaee2025llm}, and in equation generation, some works evaluate numeric equivalence or label matching against human annotations~\citep{zong2023solving}. Comprehensive rubric-based human evaluation remains insufficient, leaving gaps in assessing  generation fidelity. 

To address these gap, our evaluation framework combines multi-level automatic metrics with \ac{LLM}-based rubric scoring to capture lexical, syntactic, and semantic accuracy together with human assessment to analyze alignment, enabling a more complete understanding of generation quality.

\section{Evaluation Workflow}

\begin{figure}[t]
    \centering
\includegraphics[width=\linewidth,
        trim=20 15 20 5,
        clip]{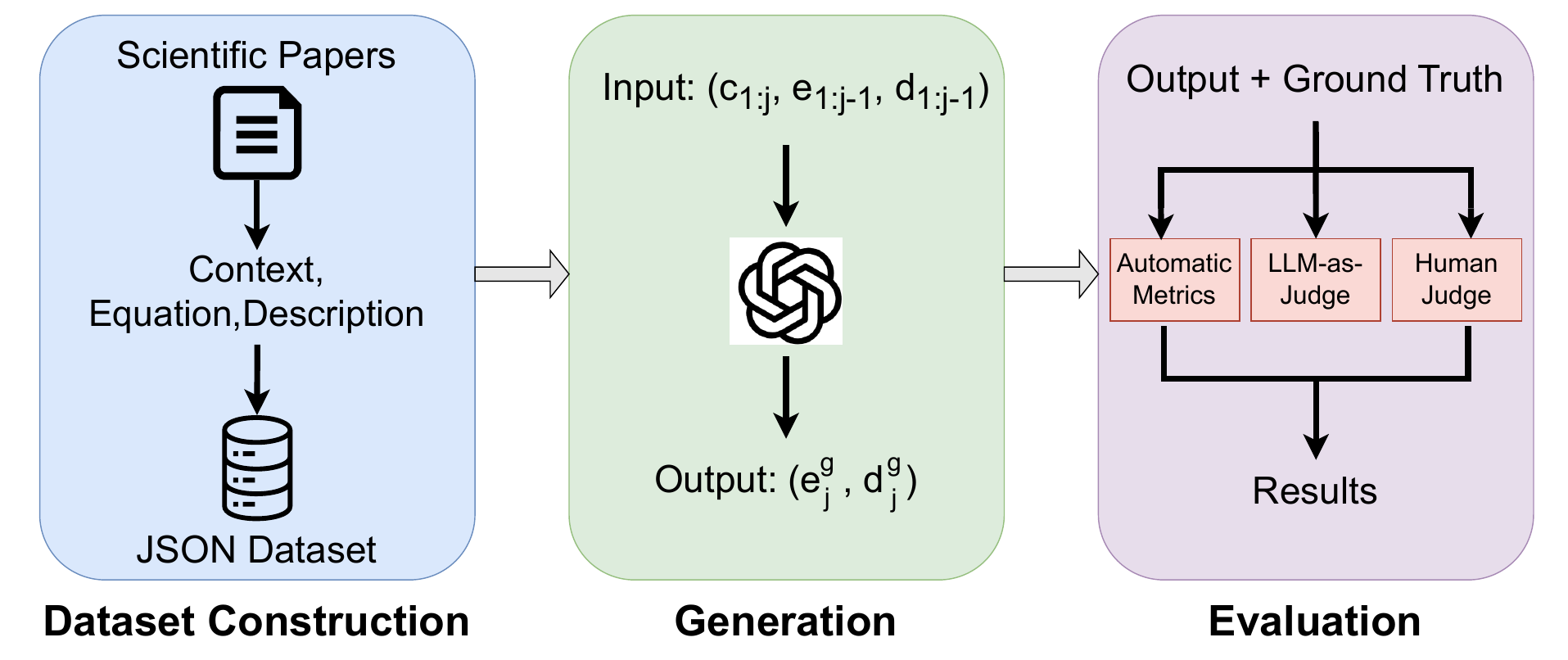}
    \caption{The proposed  workflow for explainable equation generation in
scientific discovery.} 
    \label{fig:workflow} 
\end{figure}

As shown in \autoref{fig:workflow}, our workflow has three stages:
\begin{enumerate*}
    \item \textit{Contextual Equation Corpus Creation} (\S\ref{sec:dataset}), where equations are preserved within original narrative contexts from research papers.
    \item \textit{Context-to-Equation Generation} (\S\ref{sec:generation}), where \acp{LLM} generate equations and descriptions based on the corpus.
    \item \textit{Multi-Level Equation Evaluation} (\S\ref{sec:evaluation}), assessing outputs via automatic metrics, \ac{LLM}-based rubrics, and human evaluation.
\end{enumerate*}

\subsection{Contextual Equation Corpus Creation}\label{sec:dataset}
\paragraph{Data Source.}
The dataset was constructed to comprehensively cover recent research in \acs{AI} domain, with a focus on papers containing significant mathematical or formal content. Papers were sourced from leading AI conferences archived in the ACL Anthology and OpenReview, including representative venues such as ACL and ICLR. We collected papers published in the past two years and applied selection criteria to ensure high-quality content, including methodological significance and the presence of multiple equations, ultimately including 170 papers in the dataset.

\paragraph{Dataset Construction.}
We extracted the first four pages of short papers and the first seven pages of other papers with equations.

We converted PDFs to Markdown using the META Nougat tool~\citep{blecher2024nougat}, which preserves LaTeX equations and tables, removed files with failed conversions or with fewer than 2 or more than 15 equations. We merged multiple expressions appearing in a single block into one entry to preserve the grouping of related equations. 
The data schema is provided (see Appendix~\ref{sec:dataset_structure}).

Formally, the dataset is $\mathcal{D} = \{p_1, \dots, p_n\}$, where $n$ denotes the total number of selected papers.
Each paper $p_i$ is extracted as a list of $n_i$ context-equation-description triples: $p_i=[(c_1, e_1, d_1), \dots, (c_{n_i}, e_{n_i}, d_{n_i})]$, where $c_j$ is the natural language context of the $j$-th equation, $e_j$ is the corresponding LaTeX equation, and $d_j$ is a concise description of variables in $e_j$.

\begin{table}[H]
\centering
\begin{adjustbox}{width=\columnwidth}
\begin{tabular}{lccccc}
\hline
Venue & \#Paper & Avg \#Eq & 2--5 Eqs & 6--10 Eqs & 11+ Eqs \\
\hline
ACL     & 21  & 5.3  & 12 & 6  & 3  \\
NAACL   & 28  & 5.1  & 17 & 9  & 2  \\
NeurIPS & 32  & 7.8  & 11 & 10 & 11 \\
ICLR    & 46  & 7.8  & 11 & 25 & 10 \\
ICML    & 43  & 8.1  & 14 & 18 & 11 \\
\hline
Total   & 170 & 6.1  & 65 & 68 & 37 \\
\hline
\end{tabular}
\end{adjustbox}
\caption{Equation statistics by venues.}
\label{tab:equation_stats}
\end{table}

\paragraph{Dataset Statistics.} 
    After preprocessing and filtering the collected papers, we analyzed the distribution of equations to understand typical usage patterns. We analyzed 170 papers that each contained two or more equations, totaling 1,043 equations. As shown in \autoref{tab:equation_stats}, most papers have 2 to 10 equations, while 37 papers have 11 or more. By venue, ICML, ICLR, NeurIPS papers are more equation-heavy (avg. 7.8--8.1 eqs/paper) than ACL and NAACL (avg. 5 eqs/paper), reflecting different writing styles.

\subsection{Explainable Equation Generation}\label{sec:generation}

\begin{algorithm}[t]
\tiny
\caption{Equation Generation Workflow}
\label{alg:eq_generation}
\begin{algorithmic}[1]
\Require Paper set $P = [p_1, \dots, p_{n}]$, $n_i$ is max number of equations in $i$-{th} paper.
\Statex \textbf{Initialize state:} 
\State $S.\text{equations} \gets []$ \Comment{Store generated equations and descriptions}
\State $S.\text{current\_paper} \gets 0$
\State $S.\text{current\_step} \gets 0$
\State $S.\text{finished} \gets \text{False}$
\For{$p_i \in P$}
    \State $M \gets \textsc{Extract\_method}(p_i)$ \Comment{Extract core methodology section}
    \State $\{c_1, \dots, c_{n_i}\} \gets \textsc{detect\_context}(M)$ \Comment{Original context}
    \State $\{e_1, \dots, e_{n_i}\} \gets \textsc{detect\_equation}(M)$ \Comment{Original equations}
    \State $\{d_1, \dots, d_{n_i}\} \gets \textsc{detect\_description}(M)$ \Comment{Original descriptions}
    \For{$j = 1$ to $n_i$}
        \State $\mathcal{I}_j \gets \textsc{construct\_input}(c_{1:j}, e_{1:j-1}, d_{1:j-1})$ 
        \State $(e^g_j, d^g_j) \gets \text{LLM}(\mathcal{I}_j)$ \Comment{Generate equation and description}
        \State $S.\text{equations.append}((e^g_j, d^g_j))$
        \State $S.\text{current\_step} \gets S.\text{current\_step} + 1$
    \EndFor
    \State $S.\text{current\_paper} \gets S.\text{current\_paper} + 1$ \Comment{Move to next paper}
\EndFor
\State $S.\text{finished} \gets \text{True}$
\State \Return $S.\text{equations}$
\end{algorithmic}
\end{algorithm}

We formulate a \textit{context-to-equation generation} task where an \ac{LLM} generates a mathematical equation-description pair conditioned on the paper's narrative context and previously generated outputs.
Formally, given a paper $p_i \in \mathcal{D}$, the model genrates the $j$-{th} equation-description pair,
\begin{align*}
(e_j^g, d_j^g) = \text{LLM}(c_{1:j}, e_{1:j-1}, d_{1:j-1}), \forall j \in [1, n_i], 
\end{align*}

where $c_{1:j}$ denotes all context segments up to step, $e_{1:j-1}, d_{1:j-1}$ are the previous equations and descriptions of variables.

As shown in Algorithm~\ref{alg:eq_generation}, it is formulated as a document-level, sequential generation problem: at each step \(j\), an \ac{LLM} conditions on the current textual context together with all previously introduced equations and variable descriptions, and generates the next equation-description pair \((e^g_j, d^g_j)\). The input is embedded into a standardized prompt template that enforces the required output format, instructing the model to return only the LaTeX code for the equation and a concise variable description. The full prompt and an example generation process are provided in Appendix~\ref{sec:generation_prompt}.

\subsection{Multi-Level Equation Evaluation}\label{sec:evaluation}
Our evaluation protocol focuses on three primary aspects: automatic metrics, rubric metrics, and human judgment alignment. 

\paragraph{Automatic Metrics.} 
We assess equation quality using five automatic metrics: 
\begin{enumerate*}
    \item \textbf{TexBLEU}~\cite{jung2025texbleu} evaluates n-gram similarity with GPT-2 embeddings to account for both token-level and structural overlap; 
    \item \textbf{ROUGE-L}~\cite{lin2004rouge} measures longest common subsequence alignment; 
    \item Levenshtein distance (\textbf{Levenshtein-D}) ~\cite{levenshtein1966binary,li2007normalized} captures character-level edit costs normalized by length; 
    \item Sequence similarity (\textbf{SeqSim}), also called \acl{SMR} ~\cite{rao2018characteristic} quantifies contiguous matching subsequences; to capture structural accuracy;
    \item \textbf{\Acf{TED}}~\cite{schwarz2017new} compares symbolic parse trees and averages normalized distances across equations within an entry.
\end{enumerate*}
The range of all metrics are from 0 to 1, where higher values indicate greater similarity, except TED where 0 indicates identical expressions and 1 indicates maximum difference.

\begin{table}[b!]
\centering
\begin{adjustbox}{width=\columnwidth}
\begin{tabular}{l c c c c c}
\hline
\textbf{Dimension} & Context & GT\_Eq* & GT\_Des* & Gen\_Eq & Gen\_Des \\
\hline
Validity & \ding{55} & \ding{55} & \ding{55} & \checkmark & \ding{55} \\
Coverage & \checkmark & \ding{55} & \ding{55} & \checkmark & \checkmark \\
Clarity & \checkmark & \checkmark & \checkmark & \checkmark & \checkmark \\
Appropriateness & \checkmark & \ding{55} & \ding{55} & \checkmark & \checkmark \\
Equivalence & \checkmark & \checkmark & \checkmark & \checkmark & \checkmark \\
\hline
\end{tabular}
\end{adjustbox}
\caption{Information used in each \ac{LLM}-as-Judge evaluation dimension. * indicates ground-truth; unmarked columns correspond to generated outputs.}
\label{tab:llm_judge_info_matrix}
\end{table}

\paragraph{Rubric Metrics.}\label{sec:evaluation_metric_rubric}

We define five-dimensional rubrics for LLM and human judges to evaluate both surface-level correctness and semantic and contextual alignment:
\begin{enumerate*}
    \item \textbf{Validity}, examining whether the equation is syntactically and mathematically well-formed;
    \item \textbf{Coverage}, measuring whether all essential components are included; 
    \item \textbf{Clarity}, evaluating the clarity of the generated accompanying description;
    \item \textbf{Appropriateness}, evaluating the degree to which the generated outputs appropriately match the scenario, intent, or constraints of the original problem statement;
    \item \textbf{Equivalence}, assessing whether the generated equation conveys the same mathematical meaning as the ground-truth equation, allowing for algebraically equivalent transformations and variable renaming.
\end{enumerate*}

For each rubric metric, we provide detailed instructions in Appendix~\ref{sec:rubric_metrics} and specific information to guide its evaluation in \autoref{tab:llm_judge_info_matrix}.

\begin{table*}[htb!]
\centering
\begin{adjustbox}{width=\textwidth}
\begin{tabular}{lccccccc}
\hline
\textbf{Model} & \textbf{TexBLEU (\%)} & \textbf{Levenshtein (\%)} & \textbf{SeqSim (\%)} & \textbf{ROUGE-L (\%)} & \textbf{\textbf{Lexicality (\%)}} & \textbf{TED (\%)} & \textbf{\textbf{
Syntax (\%)}} \\
\hline
GPT-4.1     & 57.4  & 45.6  & 55.1  & 57.8 & \textbf{54.0} & 69.3 & \textbf{30.7} \\
DeepSeek-R1 & 57.2 & 41.5 & 51.0 & 54.7 & \textbf{51.1} & 71.1 & \textbf{28.9} \\
GPT-4o-mini & 54.1  & 42.3 & 51.3 & 52.8 & \textbf{50.1} & 71.0  & \textbf{29.0 } \\
LLaMA3-70B  & 54.6 & 42.5  & 51.1  & 52.3  & \textbf{50.1 } & 70.7  & \textbf{29.3 } \\
Qwen3-235B  & 59.8  & 44.1  & 53.1  & 55.1  & \textbf{53.0 } & 69.9  & \textbf{30.1 } \\
\hline
Overall  & 56.6 $\pm$ 2.1 & 43.2 $\pm$ 1.5 & 52.3 $\pm$ 1.6 & 54.5 $\pm$ 2.0 & \textbf{51.7 $\pm$ 1.6}  & 70.4 $\pm$ 0.7 & \textbf{29.6 $\pm$ 0.7} \\
\hline
\end{tabular}
\end{adjustbox}
\caption{
Automatic evaluation results across lexical (TexBLEU, Levenshtein-D, SeqSim, ROUGE-L) and syntactic (1 - \ac{TED}) metrics. Lexicality is the mean of the four lexical metrics, capturing token- and sequence-level agreement, while Syntax measures hierarchical structure recovery.
}
\label{tab:metrics_summary}
\end{table*}

\paragraph{Human Judgment Alignment.}
In addition to \ac{LLM}-based evaluation, we conducted a human evaluation to examine alignment between \acp{LLM} and human judgments. A total of 100 examples from the evaluation dataset were independently assessed using the same five rubrics as the \ac{LLM}-as-Judge. The evaluation was distributed across five human judges, and their assessments were aggregated to create a combined reference for comparison. 
Human judges were provided with the same rubrics available to \acp{LLM}.
We assess the alignment by correlation coefficients:
\begin{enumerate*}[leftmargin=*, label=(\arabic*)]
  \item \textbf{Pearson's $r$}: Evaluates linear correlation between continuous variables.
  \item \textbf{Spearman's $\rho$}: Measures monotonic relationships between ranked variables. And 
  \item \textbf{Kendall's $\tau$}: Assesses ordinal associations between ranked variables.
\end{enumerate*}

\section{Experimental Setup}
\subsection{Research Questions}
This work is guided by the following research questions:
\begin{enumerate*}[leftmargin=*, label=(\textbf{RQ\arabic*}), nosep]
    \item How accurately can \acp{LLM} generate correct mathematical equation from unstructured real-world scientific literature?
    \item How does the number of equations in the original text influence the accuracy of the generation?
    \item how well do \ac{LLM} evaluation align with human assessments in generation tasks?
\end{enumerate*}

\subsection{Backbone Models}

We evaluate equation generation performance across a diverse set of \acp{LLM}, covering both general-purpose and math-oriented architectures to enable comparative analysis. Specifically, we include the following models: 
\begin{enumerate*}
    \item OpenAI GPT-4.1, 
    \item OpenAI GPT-4o-mini,
    \item DeepSeek-R1,
    \item LLaMA3.3-70B,
    \item  Qwen3-235B.
\end{enumerate*}
OpenAI~\footnote{\url{https://platform.openai.com/docs/models}} and DeepSeek~\footnote{\url{https://api.deepseek.com/models}} models are accessed via their native APIs, while LLaMA and Qwen models are queried through the Together AI platform~\footnote{\url{https://www.together.ai/}}.

\subsection{Implementation details}
The proposed workflow is implemented by Python.
During the equation generation stage, we use five representative \ac{LLM} with their respective default decoding configurations.
Specifically, Qwen3 is used with its default temperature of 0.7; GPT-4.1 and GPT-4o are used with temperature set to 1.0; and LLaMA-3 and DeepSeek-R1 are used with temperature set to 0.6. The input is formatted using a standardized prompt template (see Appendix~\ref{sec:generation_prompt}) to enforce consistent LaTeX and variable description outputs. All generated equations and descriptions are stored in structured JSON files indexed by paper and equation identifiers.
To complement automatic metrics, we use GPT-4o-mini as an evaluator with a fixed temperature of 0.2. Each evaluation prompt provides the same contextual information used during generation, ensuring reliable and easily parseable LLM scoring. Detailed evaluation prompt templates are provided in Appendix~\ref{sec:evaluation_prompt}.

\section{Main Results}\label{sec:results}

\subsection{Accuracy of Equation Generation (RQ1)}\label{sec:rq1_accuracy}
\subsubsection{Evaluation with Automatic Metrics}
\autoref{tab:metrics_summary} reports all models' equation generation performance using automatic metrics.
First, the models achieve moderate accuracy in recovering mathematical expressions from unstructured scientific text. 
The highest scores across four lexical similarity metrics remain below 60\%, indicating persistent challenges in reliably generating mathematical equation from real-world scientific writing. 
Second, \textit{Syntax} is substantially lower than \textit{Lexicality} ($\approx$ 20\% higher error), reflecting the difficulty of correctly capturing the hierarchical structure of mathematical expressions. 
Third, standard deviations across models are small for all metrics, indicating little variation between models.
These results partially answer (RQ1) by establishing quantitative upper bounds on current \ac{LLM} accuracy. While models demonstrate basic competence, their performance remains insufficient for highly precise scientific use, primarily due to syntactic errors. 

\subsubsection{Evaluation with Rubric Metrics.}
\begin{figure}[htb!]
    \centering
    \includegraphics[
        width=0.90\linewidth,
        trim=40 40 10 10,
        clip
    ]{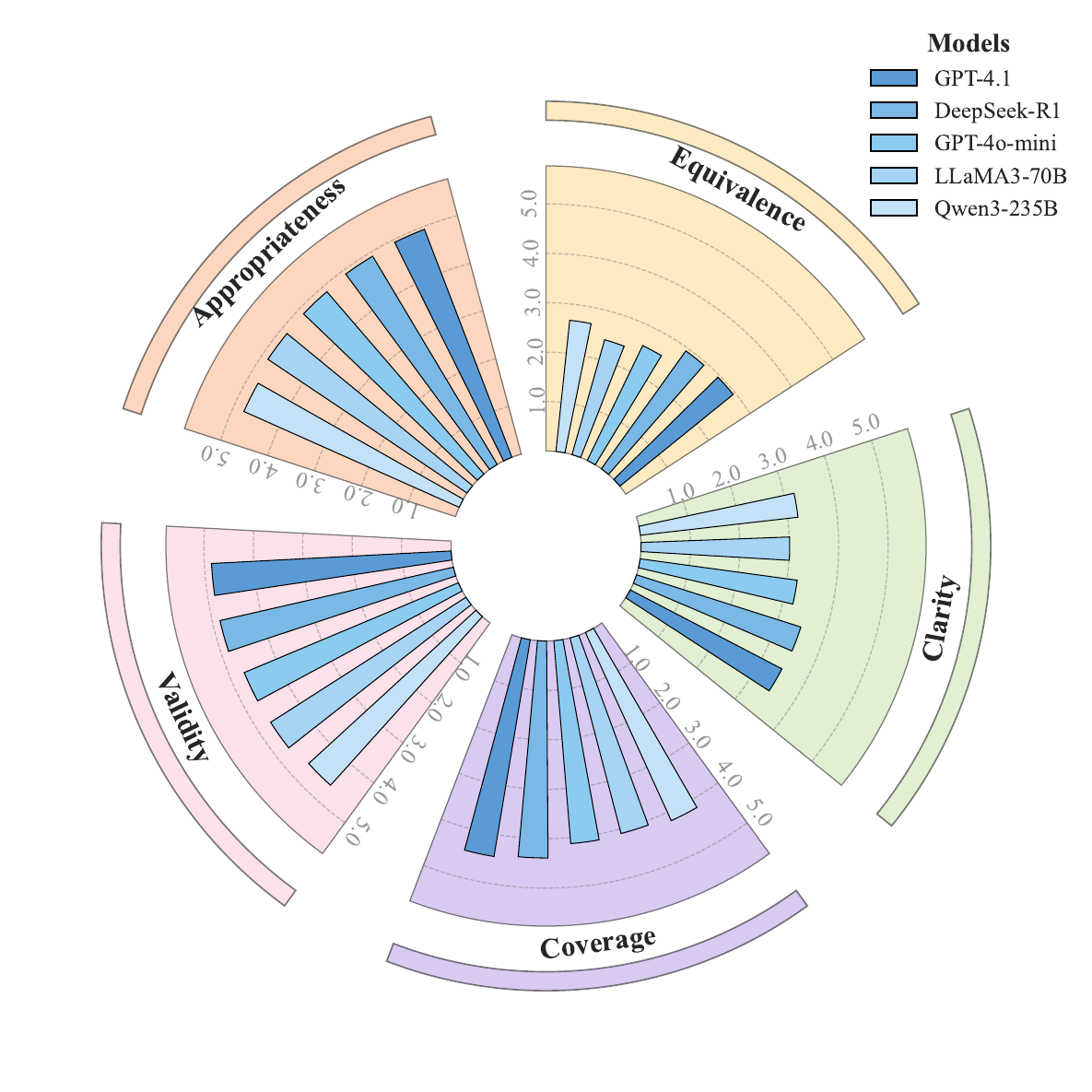}
    \caption{LLM-as-Judge average scores across metrics.}
    \label{fig:llm-scores}
\end{figure}

\autoref{fig:llm-scores} show the \ac{LLM} rubric evaluation results.

First, models generally produce well-formed equations that fit the input context, as reflected in high \textit{Validity} and contextual \textit{Appropriateness} scores around 4.7 to 4.9. However, \textit{Equivalence} remains low across all models, ranging from approximately 2.4 to 3.0, and description clarity is moderate, around 3.0 to 3.5, indicating limitations in generating logically sound outputs. 

Second, components \textit{Coverage} scores are relatively high, approximately 4.1 to 4.5, suggesting that most outputs contain sufficient components to resemble plausible answers even when they are not fully correct. 

Third, scores of \textit{Clarity} scores of the generated descriptions are relatively low, indicating that the descriptions do not clearly explain the variables in the generated equations or exhibit some inconsistencies with the original equations and descriptions.

Overall, these results show that while \acp{LLM} can generate contextually appropriate and syntactically correct formulas, their ability to faithfully reproduce the intended mathematical meaning is limited, highlighting challenges in capturing the semantics of equations and directly addressing (RQ1).

\subsection{Effect of Number of Equations (RQ2)}\label{sec:num_eq}
\begin{figure}[htb!]
    \centering
    \includegraphics[width=\linewidth]{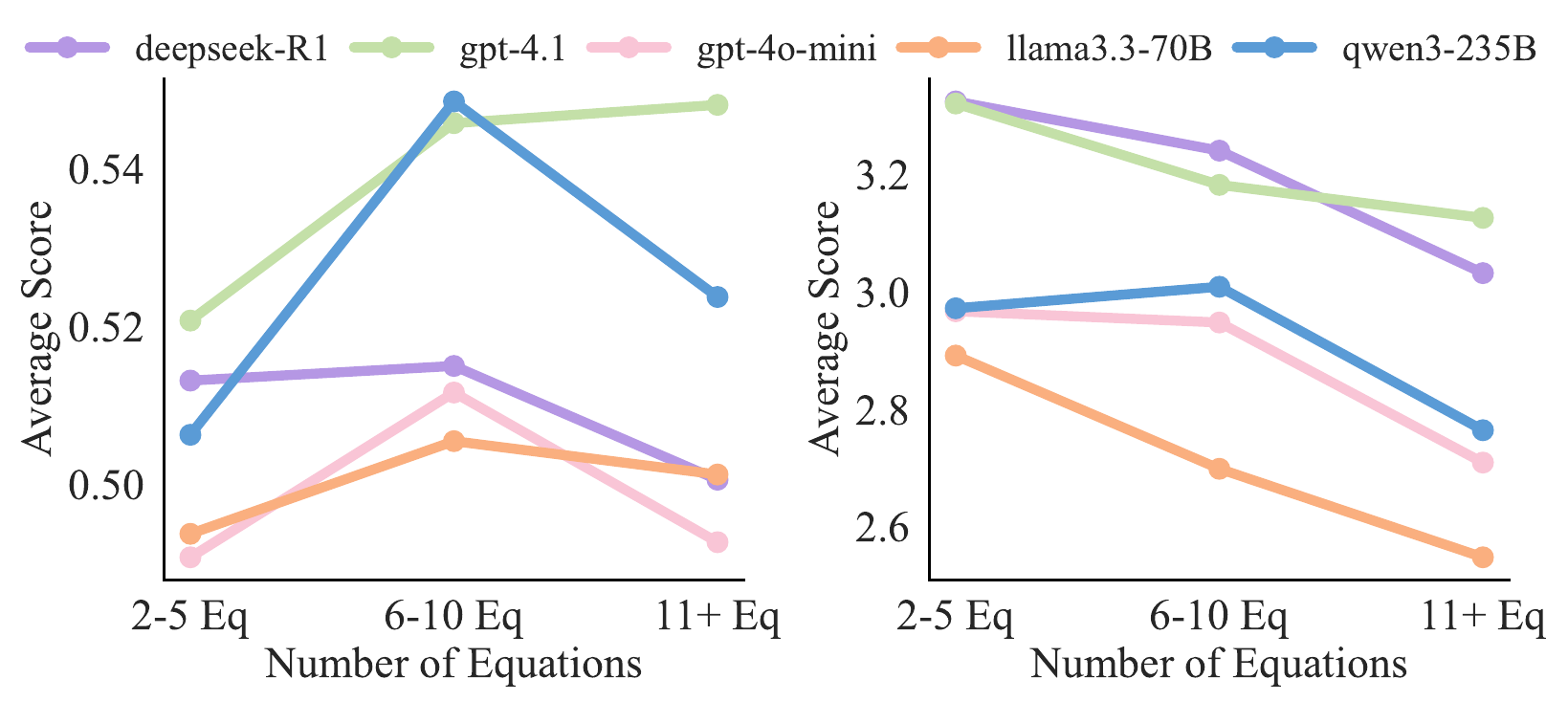}
    \vspace{1mm} 
    \begin{minipage}{0.48\linewidth}
        \centering
        \small (a) Automatic metrics.
    \end{minipage}%
    \hfill
    \begin{minipage}{0.48\linewidth}
        \centering
        \small (b) \ac{LLM} rubric metrics.
    \end{minipage}
    \caption{Effect of equation count on generation accuracy. Papers are grouped by equation numbers: 2–5, 6–10, and 11+.}
    \label{fig:number_equation}
\end{figure}

To analyze the effect of the number of equations on generation accuracy, papers are grouped into three categories based on equation count: 2--5, 6--10, and 11 or more. 
We examine the average performance using both automatic metrics and \ac{LLM}-based rubric metrics (see \autoref{fig:number_equation}). 

The results of automatic metrics show a slight increase for papers with 6--10 equations, followed by a decrease for papers with more than 10 equations. This result suggests that a moderate number of equations may provide sufficient context for models to better match the form of equations, while a larger number of equations may increase challenges that reduce lexical and syntactic accuracy. 

In contrast, \ac{LLM}-based evaluations show a clear downward trend as the number of equations increases, indicating that larger numbers of equations make it more difficult for models to maintain semantic accuracy and reasoning quality. 

These results indicate that the increase in equation numbers negatively impacts the performance of \acp{LLM}, particularly in capturing the intended meaning of mathematical expressions, and that automatic metrics may underestimate the impact of equation density on accuracy.

\begin{figure}[h!]
    \centering
    \includegraphics[
        width=0.9\linewidth,
        trim=5 10 5 5,
        clip
    ]{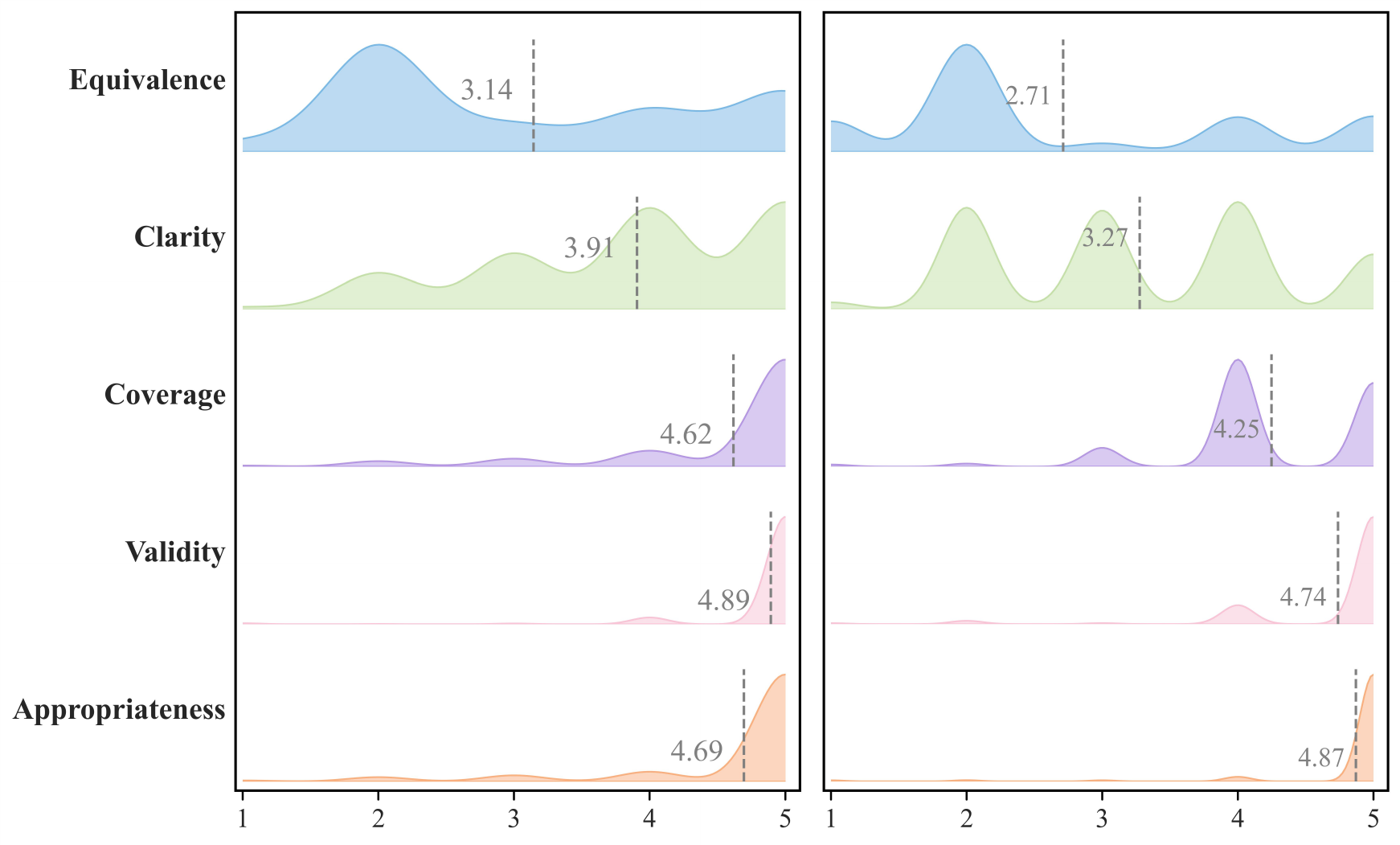}

\vspace{-0.3em}
\begin{minipage}[t]{0.50\linewidth}
    \scriptsize
    \hfill (a) Human evaluation
\end{minipage}
\hfill
\begin{minipage}[t]{0.35\linewidth}
    \scriptsize
    (b) Model evaluation
\end{minipage}
    \caption{Comparison of the distribution of human-evaluated scores and model-evaluated scores.}
    \label{fig:human_llm}
\end{figure}

\subsection{Alignment with Human Judgments (RQ3)}\label{sec:alignment}

\autoref{fig:human_llm} shows that human judges consistently assign higher scores than \ac{LLM}-as-Judge across all five evaluation metrics, including semantic accuracy, description clarity, component coverage, syntax validity, and context alignment. 

\begin{figure}[htb!]
    \centering
\includegraphics[width=\linewidth,
        trim=5 12 5 7,
        clip]{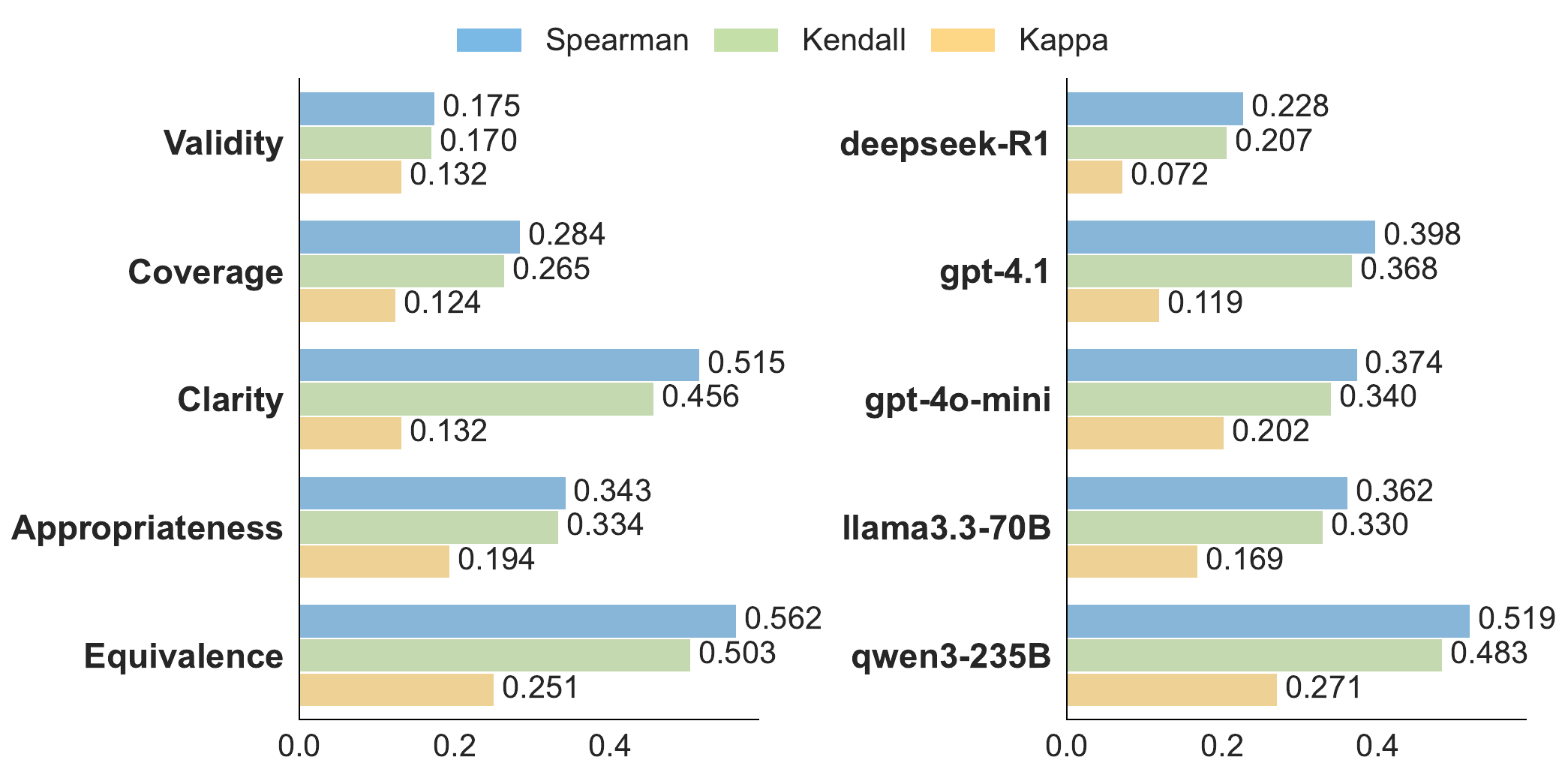}
\vspace{-0.1em}
\begin{minipage}[t]{0.40\linewidth}
    \scriptsize
    \hfill (a) Metric-level
\end{minipage}
\hfill
\begin{minipage}[t]{0.26\linewidth}
    \scriptsize
    (b) Model-level
\end{minipage}
    \caption{Alignment between LLM-based and human evaluations measured by correlation coefficients.}
    \label{fig:alignment}
\end{figure}

Correlation analysis reveals that DeepSeek-R1 achieves the highest alignment with human judgments, with moderate Spearman and Kendall's tau correlations for semantic accuracy (see \autoref{fig:alignment}). Other models and dimensions exhibit generally weak correlations, with many near zero or negative, indicating inconsistent ranking behavior compared to humans. Cohen's kappa values are uniformly low across all models, mostly around 0.2, reflecting moderate agreement in exact score assignment.

Overall, these results suggest that \ac{LLM}-as-Judge provides limited alignment with human evaluation, showing partial consistency only in certain models and dimensions, and highlights the challenges of relying on automated judgments to fully replicate human assessment, directly answering (RQ3).

\section{Additional Analysis}

\subsection{Performance Over Different Venues}

\begin{figure*}[htb!]
    \centering
\includegraphics[width=\linewidth]{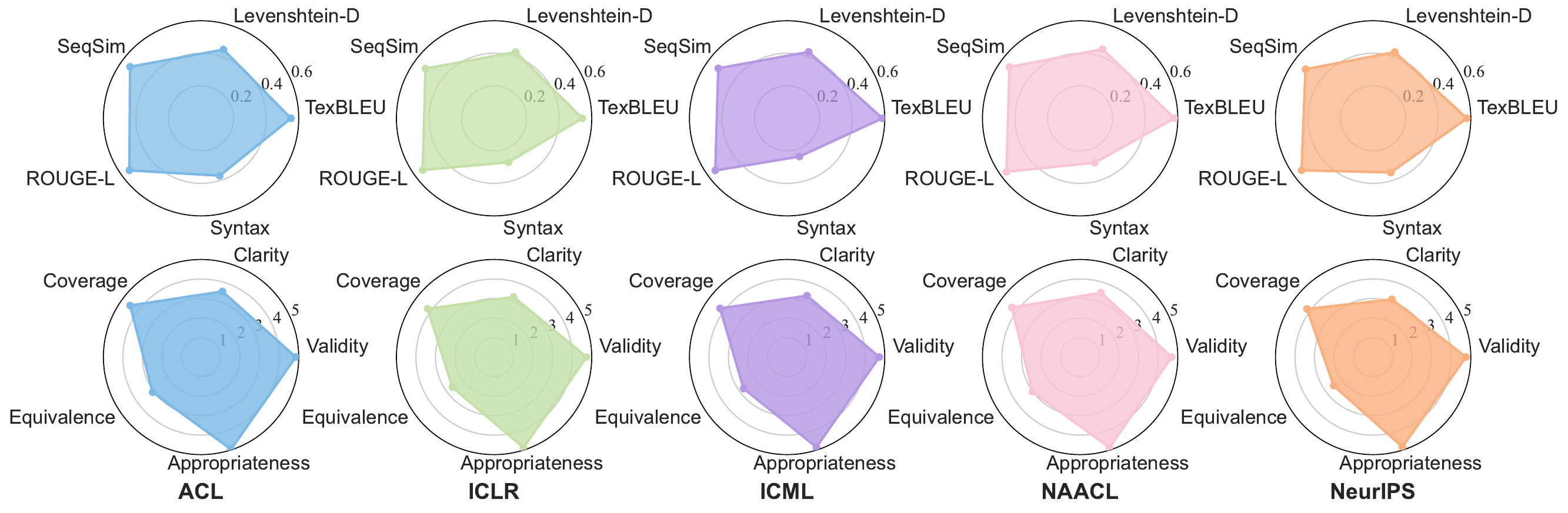}
    \caption{Scores for different venues.}
    \label{fig:venues}
\end{figure*}
\autoref{fig:venues} shows evaluation scores across five venues, using both automatic metrics and LLM-as-Judge rubric-based scores. 

First, the scores demonstrate minimal variation across venues: no conference consistently yields higher or lower values on any of the ten metrics. This result indicates that differences in writing style and equation density among these venues do not substantially affect the evaluation outcomes. 

Second, \textit{Syntax} and \textit{Equivalence} consistently receive lower scores than other metrics in all venues, highlighting the difficulty of capturing mathematical structure and symbolic equivalence beyond surface-level formula understanding and matching.

\subsection{Evaluation with Explanations}
Besides generating scores, our \ac{LLM}-based evaluation framework produces natural-language explanations for each rating. 
The explanations provide insight into the specific reasons behind a low or high rating, highlighting differences in variable usage, equation structure, or semantic interpretation. 

\begin{table}[htb!]
\centering
\small
\resizebox{\columnwidth}{!}{
\begin{tabular}{p{0.2\columnwidth} p{0.8\columnwidth}}
\hline
\textbf{Context} & Here, $ops$ is a sequence of edit operations transforming $G_1$ into $G_2$, with $w(op_i)$ representing the cost of the $i$-th operation.  \texttt{Normalization:} The tree edit distances are normalized to account for the complexity of the code by considering the maximum number of nodes between the two trees. A ramp function is added to avoid extreme values.\\
\hline
\textbf{Reference} & 
$\text{TSED}=\max\{1-\frac{\delta}{MaxNodes(G_{1},G_{2})},0\}$

 \\
\hline
\textbf{Generation} & $\text{TSED}(G_1, G_2) = \min\left( \frac{\Delta(G_1, G_2)}{\max(|G_1|, |G_2|, 1)}, 1 \right)$ \\
\textbf{Score} & 1 \\
\textbf{Explanation} & 
The generated equation fundamentally alters the relationship expressed in the ground truth by using a minimum instead of a maximum, leading to a significant misunderstanding of the intended normalization process. \\
\hline
\end{tabular}
}
\caption{Example of evaluation outcome with explanation. 
}
\label{tab:llm_explanation_example}
\end{table}

For instance, in \autoref{tab:llm_explanation_example}, the low score 1 on equivalence is caused by an incorrect normalization process:
The generated equation uses ``min'' instead of ``max'', altering the semantics of the reference formulation.
By offering interpretable feedback alongside numerical scores, this reasoning-enhanced evaluation not only increases transparency but also facilitates qualitative analysis of \ac{LLM} outputs, allowing researchers to better understand where and why generated equations deviate from the intended meaning. Overall, integrating reasoning explanations improves the utility of \ac{LLM}-as-Judge for evaluating equation generation, making the assessment more informative and actionable.

\subsection{Operator-level Equivalence Analysis}

\autoref{tab:operator-level-equivalence} shows an operator-level equivalence analysis based on Jaccard overlap~\citep{ jaccard1912distribution} to provide an objective structural signal, complementing our evaluation of semantic correctness.
\begin{table}[h!]
\centering
\begin{adjustbox}{width=\columnwidth}
\begin{tabular}{l c c c c c}
\hline
\textbf{Model} & Exact ($J=1.0$) & High ($J \geq 0.8$) & Low ($J < 0.8$) \\
\hline
DeepSeek-R1 & 20\% & 33\% & 67\% \\
Qwen3-235B & 26\% & 39\% & 61\% \\
GPT-4.1 & 24\% & 37\% & 63\% \\
GPT-4o-mini & 22\% & 39\% & 61\% \\
LLaMA3.3-70B & 20\% & 33\% & 67\% \\
\hline
\end{tabular}
\end{adjustbox}
\caption{Equivalence analysis based on Jaccard overlap over operator sets, evaluated on randomly sampling 100 generated equations per model.}
\label{tab:operator-level-equivalence}
\end{table}

Approximately 20--26\% of generated equations achieve exact operator equivalence, while 33--39\% demonstrate high operator overlap. This suggests that models partially recover the underlying mathematical structure even when strict symbolic equivalence is not achieved.
This finding is consistent with our earlier observations (\S~\ref{sec:rq1_accuracy}), where models produce syntactically valid yet semantically insufficient equations, indicating a persistent gap between structural correctness and semantic equivalence.

\section{Discussion and Implication}

\paragraph{Complementary Evaluation Fidelity.}
We combine automatic metrics, \ac{LLM}-based rubrics, and human judgments to evaluate equation accuracy, explainability, and alignment. 
Results show that single-metric evaluation is biased, with mathematical unit identification and symbolic semantics remaining key bottlenecks.
Models often misidentify core variables, causing semantic errors despite well-formed expressions, indicating a tendency to favor surface plausibility over semantic accuracy.

\paragraph{Model Capacity vs.\ Equation Density.} 
Our results provide nuanced insights into how \acp{LLM} process structured mathematical content embedded in unstructured text, revealing a clear interaction between model capacity and equation density. We observe a non-monotonic effect of equation density on generation performance. Generation performance improves slightly for papers with a moderate number of equations (6–10), indicating helpful contextual cues, but declines as density increases, especially for semantic accuracy, suggesting limitations in maintaining long-range symbolic dependencies.

\paragraph{Minimal Venue-Specific Writing Style Effects.}
Although the number of equations varies by venue -- ICML, ICLR, and NeurIPS are more equation-heavy (around eight equations per paper) than ACL and NAACL (around five) -- model performance remains largely stable across venues (\S\ref{sec:num_eq}). 
Evaluation scores are also consistent across venues (\S\ref{sec:alignment}). 
These results suggest that the observed limitations arise from inherent model capabilities rather than writing style or venue-specific conventions.
We caution against treating venue characteristics as proxies for scientific creativity or difficulty.

\paragraph{Operator-level Equivalence.}
While formal symbolic equivalence checking is a desirable direction, many equations in our setting involve context-defined operators (e.g., softmax, encoder, policy functions), whose semantics depend on the surrounding scientific context rather than canonical algebraic identities. This limits the applicability of traditional symbolic normalization tools. We therefore approximate structural equivalence by computing the Jaccard overlap between operator sets extracted from generated and reference equations.

\paragraph{Implications.}
These findings indicate the need for evaluation frameworks and modeling approaches that explicitly account for symbolic structure, inter-equation dependencies, and semantic equivalence, beyond surface-level textual similarity. SciText2Eq provides a general and extensible workflow for both modeling and evaluation in equation generation. Built from recent \acs{AI} papers, it is not domain-specific, and its methods for contextual equation corpus construction, context-to-equation generation, and multi-level evaluation can apply to other formal scientific domains such as physics, biology, economics, and engineering.

\paragraph{Future Work.}
We plan to extend the workflow to enable cross-paper reasoning by incorporating a retriever that surfaces related equations together with their original contexts from other papers. Another promising direction is to apply reinforcement learning to optimize equation generation with respect to both syntactic structure and semantic correctness. Finally, future work may improve evaluator alignment by combining rubric-based \ac{LLM} judgments with lightweight human-in-the-loop calibration, enabling more reliable evaluation of scientific equation generation.

\section{Conclusion}
In this work, we investigated the ability of \acp{LLM} to generate mathematical equations from real scientific texts. 
We introduce an explainable equation generation framework evaluated on diverse open- and closed-source LLMs, along with a unified evaluation protocol combining automatic metrics, LLM-based rubrics, and human judgments to measure accuracy, explainability, and human--LLM alignment.
Our analyses reveal that \acp{LLM} can produce syntactically valid and contextually appropriate equations. 
Meanwhile, they often struggle with semantic accuracy in scientific documents with high equation density.
By framing equation generation as a proxy for creative reasoning, we highlight how accurate extraction and reconstruction of mathematical expressions is essential for identifying, combining, and building upon conceptual relationships across studies. 
Overall, our work highlights the need for improved models and evaluation strategies to better capture the semantic and structural differences of scientific equations, paving the way toward AI-assisted scientific discovery.

\section*{Limitations}
Our work reveals several limitations of current \acp{LLM} in generating mathematical equations from scientific texts. Despite moderate performance on lexical- and syntactic-based metrics, models struggle with semantic accuracy, particularly when papers contain a higher density of equations. This gap underscores the difficulty of accurately interpreting and reconstructing mathematical content in realistic scientific contexts, beyond the simplified settings of existing benchmarks.
These limitations have direct implications for scientific creativity. Generating equations from unstructured text can be viewed as a proxy for creative scientific reasoning, requiring the ability to identify, interpret, and recombine conceptual relationships across studies. When \acp{LLM} fail to fully capture the intended meaning of equations or miss subtle dependencies, their outputs may misrepresent the underlying concepts, potentially constraining the ability of researchers to synthesize knowledge efficiently and explore novel ideas. Thus, improving semantic understanding and reasoning in equation generation is essential not only for model performance but also for supporting human scientific creativity and accelerating the research process.

\section*{Ethical Consideration}
This work focuses on evaluating and improving automated scientific equation generation and does not involve human subjects, personal data, or sensitive information, and we see no potential risks. All data are sourced from publicly available research papers and used solely for research purposes. While \acp{LLM} can generate mathematically plausible equations, they may also produce incorrect or misleading formulations; we therefore emphasize explainability and human-aligned evaluation to mitigate risks of misuse. Our evaluation framework is intended to support, not replace, expert judgment, and we encourage responsible use of these tools in scientific workflows where final validation remains with human researchers.

\bibliography{main_CR}

@article{yu2025formalmath,
  title={FormalMATH: Benchmarking formal mathematical reasoning of large language models},
  author={Yu, Zhouliang and Peng, Ruotian and Ding, Keyi and Li, Yizhe and Peng, Zhongyuan and Liu, Minghao and Zhang, Yifan and Yuan, Zheng and Xin, Huajian and Huang, Wenhao and Wen, Yandong and Zhang, Ge and Liu, Weiyang},
  journal={arXiv preprint arXiv:2505.02735},
  year={2025},
  doi={10.48550/arXiv.2505.02735},
}

@inproceedings{shojaee2024llm,
  title={Llm-sr: Scientific equation discovery via programming with large language models},
  author={Shojaee, Parshin and Meidani, Kazem and Gupta, Shashank and Farimani, Amir Barati and Reddy, Chandan K},
  booktitle={The Twelfth International Conference on Learning Representations (ICLR)},
  year={2025},
  url={https://openreview.net/forum?id=m2nmp8P5in}
}

@inproceedings{shojaee2025llm,
  title={Llm-srbench: A new benchmark for scientific equation discovery with large language models},
  author={Shojaee, Parshin and Nguyen, Ngoc-Hieu and Meidani, Kazem and Farimani, Amir Barati and Doan, Khoa D and Reddy, Chandan K},
  booktitle={Proceedings of the Twelfth International Conference on Machine Learning (ICML)},
  year={2025},
  url={https://openreview.net/forum?id=SyQPiZJVWY}
}

@inproceedings{lin2004rouge,
  title={Rouge: A package for automatic evaluation of summaries},
  author={Lin, Chin-Yew},
  booktitle={Text summarization branches out},
  pages={74--81},
  year={2004},
  url={https://aclanthology.org/W04-1013/}
}

@inproceedings{blecher2024nougat,
  title={Nougat: Neural optical understanding for academic documents},
  author={Blecher, Lukas and Cucurull, Guillem and Scialom, Thomas and Stojnic, Robert},
  booktitle={The Twelfth International Conference on Learning Representations (ICLR)},
  year={2024},
  url={https://openreview.net/forum?id=fUtxNAKpdV}
}

@inproceedings{jung2025texbleu,
  author = {Jung, Kyudan and Kim, Nam-Joon and Ryu, Hyun Gon and Hyeon, Sieun and Lee, Seung-Jun and Lee, Hyuk-Jae},
  title = {TeXBLEU: Automatic metric for evaluating LaTeX format},
  booktitle = {IEEE International Conference on Acoustics, Speech and Signal Processing (ICASSP)},
  year = {2025},
  pages = {1--5},
  doi = {10.1109/ICASSP49660.2025.10888244},
}

@article{li2007normalized,
  title={A normalized Levenshtein distance metric},
  author={Li, Yujian and Liu, Bo},
  journal={IEEE transactions on pattern analysis and machine intelligence},
  volume={29},
  number={6},
  pages={1091--1095},
  year={2007},
  doi={10.1109/TPAMI.2007.1078}
}

@article{levenshtein1966binary,
  title={Binary codes capable of correcting deletions, insertions, and reversals},
  author={Levenshtein, Vladimir I},
  journal={Soviet Physics Doklady},
  volume={10},
  number={8},
  pages={707--710},
  year={1966},
  url={https://nymity.ch/sybilhunting/pdf/Levenshtein1966a.pdf}
}

@article{rao2018characteristic,
  title={Characteristic mining of mathematical formulas from document: A comparative study on sequence matcher and levenshtein distance procedure},
  author={Rao, G. Appa and Srinivas, G. and Rao, K. Venkata and Reddy, P. V. G. D. Prasad},
  journal={International Journal of Computer Sciences and Engineering},
  volume={6},
  number={4},
  pages={400--404},
  year={2018},
  publisher={IJCSE},
  url={https://www.ijcseonline.org/pub_paper/72-IJCSE-03242(2).pdf}
}

@inproceedings{schwarz2017new,
  title={A new perspective on the tree edit distance},
  author={Schwarz, Stefan and Pawlik, Mateusz and Augsten, Nikolaus},
  booktitle={International Conference on Similarity Search and Applications},
  pages={156--170},
  year={2017},
  doi={10.1007/978-3-319-68474-1_11}
}

@article{song2025llmfe,
  title={LLM-Feynman: leveraging large language models for universal scientific formula and theory discovery},
  author={Song, Zhilong and Zhou, Qionghua and Ren, Chunjin and Ling, Chongyi and Ju, Minggang and Wang, Jinlan},
  journal={arXiv preprint arXiv:2503.06512},
  year={2025},
  doi={10.48550/arXiv.2503.06512}
}

@inproceedings{meadows2025controlling,
  title={Controlling equational reasoning in large language models with prompt interventions},
  author={Meadows, Jordan and Valentino, Marco and Freitas, Andr{\'e}},
  booktitle={Proceedings of the AAAI Conference on Artificial Intelligence (AAAI)},
  volume={39},
  pages={24858--24866},
  year={2025},
  doi = {10.1609/aaai.v39i23.34668}
}

@article{wang2025drsr,
  title={DrSR: LLM based scientific equation discovery with dual reasoning from data and experience},
  author={Wang, Runxiang and Wang, Boxiao and Li, Kai and Zhang, Yifan and Cheng, Jian},
  journal={arXiv preprint arXiv:2506.04282},
  year={2025},
  doi={10.48550/arXiv.2506.04282}
}

@article{du2024large,
  title={Large language models for automatic equation discovery of nonlinear dynamics},
  author={Du, Mengge and Chen, Yuntian and Wang, Zhongzheng and Nie, Longfeng and Zhang, Dongxiao},
  journal={Physics of Fluids},
  volume={36},
  number={9},
  year={2024},
  publisher={AIP Publishing},
  doi={10.1063/5.0224297}
}

@inproceedings{Merler2024icsr,
  title={In-context symbolic regression: Leveraging large language models for function discovery},
  author={Merler, Matteo and Haitsiukevich, Katsiaryna and Dainese, Nicola and Marttinen, Pekka},
  booktitle={Proceedings of the 62nd Annual Meeting of the Association for Computational Linguistics (ACL SRW)},
  pages={427--444},
  year={2024},
  doi={10.18653/v1/2024.acl-srw.49}
}

@book{bais2025equations,
  title={The equations: icons of knowledge},
  author={Bais, Sander},
  year={2025},
  publisher={Routledge},
  doi={10.5117/9789053567449}
}

@inproceedings{yasunaga2019topiceq,
  title={Topiceq: A joint topic and mathematical equation model for scientific texts},
  author={Yasunaga, Michihiro and Lafferty, John D},
  booktitle={Proceedings of the AAAI Conference on Artificial Intelligence (AAAI)},
  volume={33},
  pages={7394--7401},
  year={2019},
  doi={10.1609/aaai.v33i01.33017394}
}

@inproceedings{wang2021scientific,
  title={Scientific formula retrieval via tree embeddings},
  author={Wang, Zichao and Zhang, Mengxue and Baraniuk, Richard G and Lan, Andrew S},
  booktitle={2021 IEEE International Conference on Big Data (Big Data)},
  pages={1493--1503},
  year={2021},
  doi={10.1109/BigData52589.2021.9671942}
}

@article{peng2021mathbert,
  title={Mathbert: A pre-trained model for mathematical formula understanding},
  author={Peng, Shuai and Yuan, Ke and Gao, Liangcai and Tang, Zhi},
  journal={arXiv preprint arXiv:2105.00377},
  year={2021},
  doi={10.48550/arXiv.2105.00377}
}

@inproceedings{scarlatos2023tree,
  title={Tree-based representation and generation of natural and mathematical language},
  author={Scarlatos, Alexander and Lan, Andrew},
  booktitle={Proceedings of the 61st Annual Meeting of the Association for Computational Linguistics (ACL)},
  pages={3714--3730},
  year={2023},
  doi={10.18653/v1/2023.acl-long.205}
}

@article{zhou2022end,
  title={An end-to-end formula recognition method integrated attention mechanism},
  author={Zhou, Mingle and Cai, Ming and Li, Gang and Li, Min},
  journal={Mathematics},
  volume={11},
  number={1},
  pages={177},
  year={2022},
  publisher={MDPI},
  doi={10.3390/math11010177}
}

@inproceedings{vemuganti2025advancing,
  title={Advancing math formula search using diverse structural and symbolic representations},
  author={Vemuganti, Sumedh and Seiya, Ayu and Kani, Nickvash},
  booktitle={Advances in Information Retrieval: 47th European Conference on Information Retrieval (ECIR)},
  pages={116--131},
  year={2025},
  organization={Springer},
  doi={10.1007/978-3-031-88708-6_8}
}

@article{dong2025recent,
  title={Recent Advances in Symbolic Regression},
  author={Dong, Junlan and Zhong, Jinghui},
  journal={ACM Computing Surveys},
  volume={57},
  number={11},
  pages={1--37},
  year={2025},
  publisher={ACM New York, NY},
  doi={10.1145/373563}
}

@inproceedings{zhang2023expression,
  title= {An expression tree decoding strategy for mathematical equation generation},
  author={Zhang, Wenqi and Shen, Yongliang and Nong, Qingpeng and Tan, Zeqi and Ma, Yanna and Lu, Weiming},
  booktitle={Proceedings of the 2023 Conference on Empirical Methods in Natural Language Processing (EMNLP)},
  year= {2023},
  pages={439--456},
  publisher={Association for Computational Linguistics},
  doi={10.18653/v1/2023.emnlp-main.29}
}

@inproceedings{liang2023let,
  title={Let gpt be a math tutor: Teaching math word problem solvers with customized exercise generation},
  author={Liang, Zhenwen and Yu, Wenhao and Rajpurohit, Tanmay and Clark, Peter and Zhang, Xiangliang and Kalyan, Ashwin},
  booktitle={Proceedings of the 2023 Conference on Empirical Methods in Natural Language Processing (EMNLP)},
  pages={14384--14396},
  year={2023},
  doi={10.18653/v1/2023.emnlp-main.889}
}

@inproceedings{zong2023solving,
  title={Solving math word problems concerning systems of equations with gpt-3},
  author={Zong, Mingyu and Krishnamachari, Bhaskar},
  booktitle={Proceedings of the AAAI Conference on Artificial Intelligence (AAAI)},
  volume={37},
  pages={15972--15979},
  year={2023},
  doi={doi.org/10.1609/aaai.v37i13.26896}
}

@article{koza1994genetic,
  title={Genetic programming as a means for programming computers by natural selection},
  author={Koza, John R},
  journal={Statistics and computing},
  volume={4},
  number={2},
  pages={87--112},
  year={1994},
  publisher={Springer},
  doi={10.1007/BF00175355}
}

@inproceedings{petersen2021deep,
  title={Deep symbolic regression: Recovering mathematical expressions from data via risk-seeking policy gradients},
  author={Petersen, Brenden K and Landajuela, Mikel and Mundhenk, T Nathan and Santiago, Claudio P and Kim, Soo K and Kim, Joanne T},
  booktitle={International Conference on Learning Representations (ICLR)},
  year={2021},
  url={https://openreview.net/forum?id=m5Qsh0kBQG}
}

@inproceedings{amini2019mathqa,
  title={Mathqa: Towards interpretable math word problem solving with operation-based formalisms},
  author={Amini, Aida and Gabriel, Saadia and Lin, Shanchuan and Koncel-Kedziorski, Rik and Choi, Yejin and Hajishirzi, Hannaneh},
  booktitle={Proceedings of the 2019 conference of the North American chapter of the association for computational linguistics: Human language technologies (NAACL-HLT)},
  pages={2357--2367},
  year={2019},
  doi={10.18653/v1/N19-1245}
}

@inproceedings{wang2017deep,
  title={Deep neural solver for math word problems},
  author={Wang, Yan and Liu, Xiaojiang and Shi, Shuming},
  booktitle={Proceedings of the 2017 Conference on Empirical Methods in Natural Language Processing (EMNLP)},
  pages={845--854},
  year={2017},
  doi={10.18653/v1/D17-1088}
}

@inproceedings{chiang2019semantically,
  title={Semantically-aligned equation generation for solving and reasoning math word problems},
  author={Chiang, Ting-Rui and Chen, Yun-Nung},
  booktitle={Proceedings of the 2019 conference of the North American chapter of the association for computational linguistics: Human language technologies (NAACL-HLT)},
  pages={2656--2668},
  year={2019},
  doi={10.18653/v1/N19-1272}
}

@article{jaccard1912distribution,
  title={The distribution of the flora in the alpine zone. 1},
  author={Jaccard, Paul},
  journal={New phytologist},
  volume={11},
  number={2},
  pages={37--50},
  year={1912},
  publisher={Wiley Online Library}
}

\appendix

\section{Most Related Works}\label{sec:most_related_work}

\begin{table*}[t]
\centering
\small
\setlength{\tabcolsep}{2pt}
\resizebox{\textwidth}{!}{
\begin{tabular}{p{2cm} p{2.3cm} p{2cm} p{2.8cm} p{3cm} p{3cm}}
\toprule
\textbf{Method} & \textbf{Domain} & \textbf{\#Sample} & \textbf{Input} & \textbf{Output} & \textbf{Core Metric} \\
\midrule
DrSR & Physics, Chemistry, Biology, Materials Science, etc. & 4000 equations & 
\raggedright • Variable pairs \newline • Scientific priors & 
\raggedright • Symbolic equation & 
• ACC$_\tau$/NMSE \\

LLM4ED & Physics & 22 tasks & 
\raggedright • Variable pairs \newline • Symbol library & 
\raggedright • Symbolic equation & 
• NRMSE \\

LLM-Feynman & Physics, Materials Science & 120 equations + 4 tasks & 
\raggedright • Variable pairs \newline • Scientific priors & 
\raggedright • Symbolic equations \newline • Complexity \newline • Interpretability score & 
• MAE \newline  • Interpretability Score*\\
LLM-SRBench & Physics, Chemistry, Biology, Materials Science & 239 equations & 
\raggedright • Variable pairs \newline • Problem description \newline • Scientific priors & 
\raggedright • Symbolic equations & 
• ACC$_\tau$/NMSE \newline • Symbolic Accuracy* \\

LLM-SR & Physics, Microbiology/Biology, Materials Science, etc. & - & 
\raggedright • Variable pairs \newline • Problem description \newline • Scientific priors & 
\raggedright • Equation skeleton \newline • Optimized parameters & 
• NMSE \\

ICSR & General symbolic regression & - & 
\raggedright • Variable pairs & 
\raggedright • Symbolic equation \newline • Optimized parameters & 
• R$^2$/Complexity \\

TopicEq & physics, computer science, astronomy, etc & 400k equations & 
\raggedright • Equations \newline • Surrounding text& 
\raggedright • Symbolic equation \newline • Inferred topics & 
• Topic Coherence \newline • Perplexity \newline • Syntax Error Rate \\
\midrule
SciText2Eq & \acs{AI} & 1043 equations & 
\raggedright • Context \newline • Variable description \newline • Piror equations & 
\raggedright • Symbolic equation \newline • Generated description & 
• Topic Coherence \newline • Perplexity \newline • Syntax Error Rate \\

\bottomrule
\end{tabular}
}
\caption{Comparison of prior symbolic regression / scientific equation generation works. Metrics marked with * employ LLM as a judge for evaluation.}
\label{tab:sr_comparison}
\end{table*}

\autoref{tab:sr_comparison} summarizes the most relevant prior work in equation discovery and equation generation. Our proposed workflow, \textbf{SciText2Eq}, is also listed for comparison,, highlighting its focus on context-aware generation and multi-level evaluation in the AI domain.

\section{AI Usage Disclosure}\label{sec:usage_disclosure}
\acs{AI} tools were used only in a limited capacity to assist with language editing. Specifically, we use GPT-5.2 to improve the clarity and readability of the manuscript. All scientific contributions, including research design, data collection, analysis, results, and conclusions, have been independently conducted and verified by the authors.

\section{Dataset Structure}\label{sec:dataset_structure}

We organized our dataset in a structured JSON format, capturing the hierarchical relationships between papers, equations, and their contextual information. Each entry represents a single paper, identified by a unique ID derived from its filename.

Our extraction process is \textbf{section-aware}: we drew context from the methods section to maintain semantic coherence across equations, and we systematically cleaned equations to remove extraneous or presentation-only LaTeX formatting. In addition, variable definitions were explicitly captured to preserve explanatory detail and improve interpretability.

To represent mathematical content in a structured manner, we grouped related equations into \textbf{equation groups}. Each group is associated with a unique equation ID and includes the preceding contextual text, variable descriptions (typically sentences starting with "where"), and an array of LaTeX-formatted equations. This design enables faithful representation of cases where multiple equations share a common context, such as systems of equations or step-by-step derivations. Finally, all extracted equations and their associated metadata were serialized into a structured JSON file, with an example entry shown below:

\vspace{0.5em}
\begin{lstlisting}[basicstyle=\ttfamily\small]
{
  "id": "2024.acl-short.15",
  "title": "Code-Switching Can be Better Aligners",
  "equations": [
    {
      "equation id": "1",
      "context": "Describes the joint SLU model for intent ...",
      "description": "where f(·) is the joint model ...",
      "EQ_latex": ["(o^{I}, o^{S}) = f(x)"]
    },
    {
      "equation id": "2",
      "context": "Introduces optimal transport (OT) for aligning ...",
      "description": "where norm(·) denotes row normalization ...",
      "EQ_latex": ["\\hat{Q}_{[i,j]} = \\mathrm{norm}(Q_{[i,j]})"]
    }
  ]
}
\end{lstlisting}

\section{Generation Prompt Templates}\label{sec:generation_prompt}

We used a standardized prompt template to guide equation generation from scientific text.
\autoref{tab:equation_generation_prompt} presents the generation prompt used across all experiments.
All generation prompts share the same overall structure and output format, and differ only in the inputs.

\begin{table}[!ht]
\centering
\small
\setlength{\itemsep}{0pt}
\setlength{\parsep}{0pt}
\setlength{\parskip}{0pt}
\setlength{\topsep}{0pt}

\begin{tabular}{p{\linewidth}}
\hrule 
\vspace{0.3em}
\textbf{System Prompt} \\[0.3em]

You are a scientific writing assistant trained to generate LaTeX equations for research papers.
For each equation (Equation $n$), you will receive a combined context which includes:
\begin{itemize}
    \item The surrounding natural language context for each equation up to Equation $n$.
    \item Any prior equations (Equation 1 to Equation $n-1$) along with their LaTeX representations, if $n>1$.
    \item Descriptions for prior equations (Equation 1 to Equation $n-1$), if available.
\end{itemize}

\textbf{Your task:}
\begin{enumerate}
    \item Understand the context and flow of the document.
    \item Generate the LaTeX code for the next equation (Equation $n$).
\end{enumerate}
Respond with the LaTeX code and the description only.  
\hrule
\vspace{0.3em}
\textbf{Example}\\
\textbf{Input:} Traditional works rely on text frequencies to define whether an instance is long-tail or not; thus,
low-frequency texts tend to be classified into long-tail classes. For \acp{LLM}, computing text
frequencies of previously unknown user queries is non-trivial. Following prior work
(Aimar et al., 2023; Zhong et al., 2021; Xu et al., 2021), \emph{Expected Calibration Error (ECE)}
is used to quantify ``long-tailness''. ECE measures how well predicted probabilities align with
observed accuracies (Guo et al., 2017). Formally, ECE is defined as:

\textbf{Output:}
\begin{itemize}
    \item \texttt{<latex> $\mathrm{ECE}=\sum_{i=1}^B \frac{n_{b_i}}{N} |acc(b_i)-conf(b_i)|$ </latex>}
    \item \texttt{<description> where $i$ denotes the $i$-th bin, $N$ is the total number of instances,
    $acc(b_i)$ and $conf(b_i)$ are the accuracy and confidence of bin $b_i$,
    and $n_{b_i}$ is the number of instances in bin $b_i$ </description>}
\end{itemize}
\hrule 
\end{tabular}
\caption{System and user prompt template for equation generation.}
\label{tab:equation_generation_prompt}
\end{table}

\section{Evaluation Prompt Templates}\label{sec:evaluation_prompt}

 We adopt an LLM-as-a-judge framework to evaluate generated equations and their accompanying descriptions. \autoref{tab:evaluation_prompt_table} presents the evaluation prompt used across all rubric metrics. All evaluation prompts share the same overall structure and output format, and differ only in the evaluation focus (see Appendix~\ref{sec:rubric_metrics}) and required input information (see \autoref{tab:llm_judge_info_matrix}).
 
\begin{table}[!t]
\centering
\small
\renewcommand{\arraystretch}{1.0}
\setlength{\itemsep}{0pt}
\setlength{\parsep}{0pt}
\setlength{\parskip}{0pt}
\setlength{\topsep}{0pt}

\begin{tabular}{p{\linewidth}}
\hrule
\vspace{0.3em}

\textbf{System Prompt} \\

You are a mathematical evaluation assistant trained to score generated equations and variable descriptions. For each evaluation dimension, you will receive: 
\begin{itemize}
    \item The problem context.
    \item The ground truth equations and their descriptions.
    \item The generated equations and their descriptions.
\end{itemize}

\textbf{Your task:} 
\begin{enumerate}
    \item Provide a score from 1 to 5 according to the specific evaluation dimension.
    \item Provide a brief explanation. Respond \textbf{only} in the following format:
    \begin{itemize}
        \item Score: <1-5>
        \item Explanation: <brief justification>
    \end{itemize}
\end{enumerate}

\hrule
\vspace{0.3em}
\textbf{Evaluation Input Template} \\

Evaluation Metric: Equivalence 

\begin{itemize}
    \item \textbf{Context:} \{context\} 
    \item \textbf{Ground Truth Equation(s):} \{eq\_gt\} 
    \item \textbf{Generated Equation(s):} \{eq\_gen\} 
    \item \textbf{Ground Truth Description(s):} \{description\_gt\} 
    \item \textbf{Generated Description(s):} \{description\_gen\} 
\end{itemize}

\textbf{Task:} Evaluate whether the generated equation expresses the same mathematical relationships as the ground truth, allowing for equivalent rearrangements or variable renaming. 

\textbf{Example Output:}

\begin{itemize}
    \item Score: 5
    \item Explanation: The generated equation exactly matches the ground truth in mathematical meaning, with only trivial rearrangements.
\end{itemize}

\hrule
\end{tabular}

\caption{Evaluation prompt with Equivalence metric as an example.}
\label{tab:evaluation_prompt_table}
\end{table}

\section{Rubric Metric details}
\label{sec:rubric_metrics}

As shown in \autoref{tab:rubric_metrics}, we evaluate the generated equations and descriptions using five rubric metrics. Each metric is rated on a 1--5 scale, with 5 being the best. 

\begin{table}[!h]
\centering
\small
\renewcommand{\arraystretch}{1.3}
\begin{tabular}{p{\linewidth}}
\hline
\textbf{Metric, Description and Scores} \\
\hline
\textbf{Validity} \\
Measures if equations are well-formed LaTeX, independent of correctness. \\
• 5: Fully valid; no syntax issues. \\
• 4: Minor syntax issues; easily correctable. \\
• 3: Noticeable formatting issues; still parseable. \\
• 2: Multiple syntax errors hindering understanding. \\
• 1: Completely ill-formed; not parseable. \\
\hline
\textbf{Coverage} \\
Checks if equations and descriptions provide a full solution considering intermediate steps. \\
• 5: All necessary terms, variables, and constraints present. \\
• 4: Minor omissions; still practically usable. \\
• 3: Noticeable omissions; partially interpretable. \\
• 2: Several important components missing or ambiguous. \\
• 1: Incomplete or disconnected from solution. \\
\hline
\textbf{Clarity} \\
Evaluates logical clarity and correctness of inferred relationships and generate description. \\
• 5: Clear and fully logical inferred reasoning and description. \\
• 4: Generally logical; minor ambiguities. \\
• 3: Partially clear; noticeable gaps or ambiguity. \\
• 2: Significant inconsistencies; barely understandable. \\
• 1: No coherent reasoning; confusing or nonsensical. \\
\hline
\textbf{Appropriateness} \\
Assesses whether equations and descriptions match the scenario, intent, and constraints. \\
• 5: Perfectly matches and clearly addresses context. \\
• 4: Strong alignment; minor ambiguities. \\
• 3: Partial relevance; some generic or inferred parts. \\
• 2: Loosely related; insufficient follow-through. \\
• 1: Completely irrelevant or hallucinated content. \\
\hline
\textbf{Equivalence} \\
Measures whether the generated equation expresses the same mathematical relationships as ground truth. \\
• 5: Exact same meaning; only trivial variations (algebraic rearrangement, variable renaming). \\
• 4: Near-match; minor semantic deviations, intent clearly preserved. \\
• 3: Core meaning largely correct but includes secondary inaccuracies. \\
• 2: Partial overlap; significant misunderstanding or contradiction. \\
• 1: Completely unrelated; wrong understanding. \\
\hline
\end{tabular}
\caption{Rubric metrics for \acp{LLM} and human evaluation. Each metric is listed with its description and scoring criteria.}
\label{tab:rubric_metrics}
\end{table}

\end{document}